\documentclass[11pt]{article}

\usepackage[final]{acl}

\usepackage[T1]{fontenc}

\usepackage[utf8]{inputenc}

\usepackage{microtype}

\usepackage{inconsolata}

\usepackage{tabularx}
\usepackage{xcolor}
\usepackage{graphicx}
\usepackage{xspace}
\usepackage{xcolor, colortbl}
\usepackage{array}
\usepackage{booktabs}
\usepackage{enumitem}
\usepackage{graphicx}
\usepackage{hyperref}
\usepackage{multirow}
\usepackage{multicol}
\usepackage{amsfonts}
\usepackage{amssymb}
\usepackage{times}
\usepackage{latexsym}

\newcommand\blfootnote[1]{%
  \begingroup
  \renewcommand\thefootnote{}\footnote{#1}%
  \addtocounter{footnote}{-1}%
  \endgroup
}

\usepackage{ifthen}
\newboolean{arXivVersion}
\setboolean{arXivVersion}{true}

\usepackage[toc,page,header]{appendix}
\usepackage{minitoc}

\newcommand{\llms}{LLMs\xspace}
\newcommand{\model}{\textsc{AgentReview}\xspace}

\title{\model: Exploring Peer Review Dynamics with LLM Agents}

\author{Yiqiao Jin$^{1*}$, Qinlin Zhao$^{2*}$, Yiyang Wang$^{1}$, Hao Chen$^{3}$, \\ 
\textbf{Kaijie Zhu$^{4}$, Yijia Xiao$^{5}$, Jindong Wang$^{6}$} \\
  $^{1}$Georgia Institute of Technology, 
  $^{2}$University of Science and Technology of China, \\
  $^{3}$Carnegie Mellon University, $^{4}$University of California, Santa Barbara, \\
  $^{5}$University of California, Los Angeles, $^{6}$William \& Mary\\ 
  $^{1}$\texttt{\{yjin328,ywang3420\}@gatech.edu} 
  $^{2}$\texttt{ac99@mail.ustc.edu.cn} 
  \\
  $^{3}$\texttt{haoc3@andrew.cmu.edu} $^{4}$\texttt{kaijiezhu@ucsb.edu} \\
  $^{5}$\texttt{yijia.xiao@cs.ucla.edu} 
  $^{6}$\texttt{jwang80@wm.edu} \\
  \url{https://agentreview.github.io/}
  }

\begin{document}
\doparttoc 


\maketitle
\begin{abstract}
Peer review is fundamental to the integrity and advancement of scientific publication. 
Traditional methods of peer review analyses often rely on exploration and statistics of existing peer review data, which do not adequately address the multivariate nature of the process, account for the latent variables, and are further constrained by privacy concerns due to the sensitive nature of the data. 
We introduce \model, the first large language model (LLM) based peer review simulation framework, which effectively disentangles the impacts of multiple latent factors and addresses the privacy issue. 
Our study reveals significant insights, including a notable 37.1\% variation in paper decisions due to reviewers' biases, supported by sociological theories such as the social influence theory, altruism fatigue, and authority bias. 
We believe that this study could offer valuable insights to improve the design of peer review mechanisms. Our code is available at \url{https://github.com/Ahren09/AgentReview}. 
\end{abstract}

\blfootnote{$^\ast$ Both authors contributed equally.}

\ifthenelse{\boolean{arXivVersion}}
{
}
{
\vspace{-8mm}
}

\section{Introduction}
Peer review is a cornerstone for academic publishing, ensuring that accepted manuscripts meet the novelty, accuracy, and significance standards.
Despite its importance, peer reviews often face several challenges, such as biases~\cite{stelmakh2021prior}, variable review quality~\cite{stelmakh2021prior}, unclear reviewer motives~\cite{zhang2022investigating}, and imperfect review mechanism~\cite{fox2023double}, exacerbated by the ever-growing number of submissions. 
The rise of open science and preprint platforms has further complicated these systems, which may disclose author identities under double-blind policies~\cite{sun2022does}. 

Efforts to mitigate these problems have focused on enhancing fairness~\cite{zhang2022investigating}, reducing biases among novice reviewers~\cite{stelmakh2021prior}, calibrating noisy peer review ratings~\cite{lu2024calibrating}, and refining mechanisms for paper assignment and reviewer expertise matching~\cite{xu2024one, liu2023shackles}. 
However, several challenges persist in systematically exploring factors influencing peer review outcomes: 
1) \emph{Multivariate Nature.} The peer review process is affected by a variety of factors, ranging from reviewer expertise, area chair involvement, to the review mechanism design. This complexity makes it difficult to isolate specific factors that impact the review quality and outcomes; 
2) \emph{Latent Variables.} Factors such as reviewer biases and intentions are difficult to measure but have significant effects on the review process, often leading to less predictable outcomes; 
3) \emph{Privacy Concerns.} Peer review data are inherently sensitive and carry the risk of revealing reviewer identities. Investigation of such data not only poses ethical concerns but also deters future reviewer participation. 


\begin{figure*}[t]
    \centering
    \includegraphics[width=0.99\linewidth]{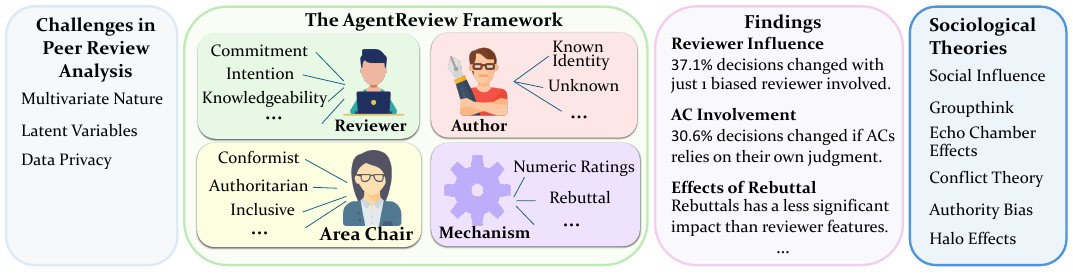}
    \ifthenelse{\boolean{arXivVersion}}
    {
    }
    {\vspace{-3mm}}
    \caption{
    \model is an open and flexible framework designed to realistically simulate the peer review process. It enables controlled experiments to \emph{disentangle} multiple variables in peer review, allowing for an in-depth examination of their effects on review outcomes. Our findings align with established sociological theories. 
    }
    \label{fig:AgentReview}
    \ifthenelse{\boolean{arXivVersion}}
    {
    }
    {
    \vspace{-2mm}
    }
\end{figure*}

\noindent \textbf{This Work.} 
We introduce \model, the first framework that integrates large language models (LLMs)~\cite{openai2023gpt4, touvron2023llama} with agent-based modeling~\cite{AutoGPT} to simulate the peer review process (Sec.~\ref{sec:framework}). 
As shown in Figure~\ref{fig:AgentReview}, \model is built upon the capabilities of LLMs to perform realistic simulations of societal environments~\cite{wu2023autogen,chen2024can, park2023generative} and provide high-quality feedback on academic literature comparable to or exceeds human levels~\cite{chen2024dolarge,chen2024drAcademy,li2024exploring,d2024marg,zhang2024comprehensive,du2024llms}. 

\model is open and flexible, designed to capture the \emph{multivariate nature} of the peer review process. It features a range of customizable variables, such as characteristics of reviewers, authors, area chairs (ACs), as well as the reviewing mechanisms (Sec.~\ref{sec:framework_overview}). 
This adaptability allows for the systematic exploration and \emph{disentanglement} of the distinct roles and influences of the various parties involved in the peer review process. 
Moreover, \model supports the exploration of alternative reviewer characteristics and more complex review processes. 
By simulating peer review activities with over 53,800 generated peer review documents, including over 10,000 reviews, on over 500 submissions across four years of ICLR, \model achieves statistically significant insights without needing real-world reviewer data, thereby maintaining reviewer \emph{privacy}.
\model also supports the extension to alternative reviewer characteristics and more complicated reviewing processes. 
We conduct both content-level and numerical analyses after running large-scale simulations of the peer review process.

\noindent \textbf{Key findings.} Our findings are as follows, which could inspire future design of peer review systems:
\begin{itemize}[leftmargin=1em]
\setlength\itemsep{0em}
    \item \textbf{Social Influence}~\cite{turner1991social}. Reviewers often adjust their ratings after rebuttals to align with their peers, driven by the pressure to conform to the perceived majority opinion. This conformity results in a 27.2\% decrease in the standard deviation of ratings (Sec.~\ref{sec:overview});
    \item \textbf{Altruism Fatigue and Peer Effects} ~\cite{angrist2014perils}. Even \emph{one} under-committed reviewer can lead to a pronounced decline of commitment (18.7\%) among all reviewers (Sec.~\ref{sec:reviewer_commitment}); 
    \item \textbf{Groupthink and Echo Chamber Effects}~\cite{janis2008groupthink, cinelli2021echo}. Biased reviewers tend to amplify each other's negative opinions through interactions (Sec.~\ref{sec:reviewer_intention}). This can lead to a 0.17 drop in ratings among biased reviewers and cause a \emph{spillover effect}, influencing the judgments of unbiased reviewers and leading to a 0.25 decrease in ratings; 
    \item \textbf{Authority Bias and Halo Effects}~\cite{nisbett1977halo}. Reviewers tend to perceive manuscripts from renowned authors as more accurate. When all reviewers know the author identities for only 10\% of the papers, decisions can change by a significant 27.7\%  (Sec.~\ref{sec:author_anonymity});
    \item \textbf{Anchoring Bias}~\cite{nourani2021anchoring}.
    The rebuttal phase, despite its role in addressing reviewers' concerns, exerts a less significant effect on final outcomes. 
    This is potentially due to anchoring bias in which reviewers rely heavily on initial impressions of the submission. 

\end{itemize}

\noindent \textbf{Contributions.} Our contributions are three-fold:
\begin{itemize}[leftmargin=1em]
\setlength\itemsep{0em}
    \item \emph{Versatile framework.} 
    \model is the first framework to employ LLM agents to simulate the entire peer review process; 
    \item \emph{Comprehensive Dataset.} We curated a large-scale dataset through our simulation, encompassing more than 53,800 generated reviews, rebuttals, updated reviews, meta-reviews, and final decisions, which can support future research on analyzing the academic peer review process;  
    \item \emph{Novel Insights.}  
    Our study uncovers several significant findings that align with sociological theories to support future research; 
\end{itemize}

\begin{figure*}[t]
    \centering
    \includegraphics[width=0.99\linewidth]{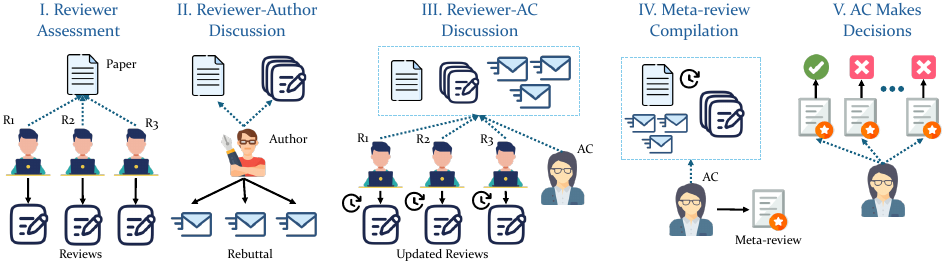}
    \ifthenelse{\boolean{arXivVersion}}
    {
    }
    {\vspace{-3mm}}
    \caption{
    Our paper review pipeline consists of 5 phases. Solid \textbf{black} arrows $\rightarrow$  represent authorship connections, while \textbf{\textcolor[HTML]{156082}{blue}} 
    dashed arrow \textcolor[HTML]{156082}{$\rightarrow$} indicate visibility relations. 
    }
    \label{fig:review_pipeline}
    \ifthenelse{\boolean{arXivVersion}}
    {
    }
    {\vspace{-4mm}}
\end{figure*}

\section{The \model Framework}
\label{sec:framework}
\subsection{Framework Overview}
\label{sec:framework_overview}

\model is designed as an extensible testbed to study the impact of various stakeholders and mechanism designs on peer review results. 
It follows procedures of popular Natural Language Processing (NLP) and Machine Learning (ML) conferences, where reviewers provide initial paper reviews, update their reviews based on authors' feedback, and area chairs (ACs) organize discussions among reviewers and make final decisions.\footnote{Some conferences or journals may follow slightly different review processes. } 
\model integrates three roles---reviewers, authors, and ACs---all powered by LLM agents. 

\noindent \textbf{Reviewers} play a pivotal role in peer review. We identify three key dimensions that determine the quality of their reviews. 
1) \emph{Commitment} refers to the reviewer's dedication and sense of responsibility in engaging with the manuscript. This involves a proactive and careful approach to provide thorough and constructive feedback on submissions.
2) \emph{Intention} describes the motivation behind the reviews, focusing on whether the reviewer aims to genuinely help authors improve their papers or is influenced by biases or conflict of interests. 
3) \emph{Knowledgeability} measures the reviewer's expertise in the manuscript's subject area.  
Understanding the effects of each individual aspect is crucial for improving the peer review process. 

To explore these dimensionalities, we assign reviewers into specific categories: knowledgeable versus unknowledgeable reviewers for \emph{knowledgeability}, responsible versus irresponsible for \emph{commitment}, and benign versus malicious for \emph{intention}. 
These categorizations are set by prompts and fed into our system as fixed characteristics. 
For example, knowledgeable reviewers are described as reviewers that are adept at identifying the significance of the research and pinpointing any technical issues that require attention. In contrast, unknowledgeable reviewers lack expertise and may overlook critical flaws or misinterpret the contributions. Reviewer descriptions and prompts are detailed in Appendix Figure~\ref{fig:prompts}. 

\noindent \textbf{Authors} submit papers to the conference and provide rebuttals to the initial reviews during the Reviewer-AC discussion period (Phase 2 in Figure~\ref{fig:AgentReview}). Although double-blind review policies are typically in place, authors may still opt to release preprints or publicize their works on social media, potentially revealing their identities. We consider two scenarios: 
1) reviewers are aware of the authors' identities due to the public release of their works, and 
2) author identities remain unknown to the reviewers. This allows us to explore the implications of anonymity on the review process. 

\noindent \textbf{Area Chairs (ACs)} have multiple duties, ranging from facilitating reviewer discussions, synthesizing feedback into meta-reviews, and making final decisions. 
ACs ensure the integrity of the review outcomes by maintaining constructive dialogues, integrating diverse viewpoints, and assessing papers for quality, originality, and relevance. 
Our work identifies three styles of ACs based on their involvement strategies, each influencing the review process differently: 
1) \emph{authoritarian} ACs dominate the decision-making, prioritizing their own evaluations over the collective input from reviewers; 
2) \emph{conformist} ACs rely heavily on other reviewers' evaluations, minimizing the influence of their own expertise; 
3) \emph{inclusive} ACs consider all available discussion and feedback, including reviews, author rebuttals, and reviewer comments, along with their expertise, to make well-rounded final decisions. 

\subsection{Review Process Design}
\label{sec:review_process_design}
\model uses a structured, 5-phase pipeline (Figure~\ref{fig:AgentReview}) to simulate the peer review process. 

\noindent \textbf{I. Reviewer Assessment.}
In this phase, three reviewers critically evaluate the manuscript. 
To simulate an unbiased review process, each reviewer has access only to the manuscript and their own assessment, preventing any cross-influence among reviewers. 
Following~\citet{liang2023can}, we ask LLM agents to generate reviews that comprise four sections, including \emph{significance and novelty}, \emph{potential reasons for acceptance}, \emph{potential reasons for rejection}, and \emph{suggestions for improvement}. This format is aligned with the conventional review structures of major ML/NLP conferences. Unless specified otherwise, each reviewer provides a numerical rating from 1 to 10 for each paper.


\noindent \textbf{II. Author-Reviewer Discussion.} 
Authors respond to each review with a rebuttal document to address misunderstandings, justify their methodologies, and acknowledge valid critiques. 

\noindent \textbf{III. Reviewer-AC Discussion.} 
The AC initiates a discussion among the reviewers, asking them to reconsider their initial ratings, and provide an updated review after considering the rebuttals. 

\noindent \textbf{IV. Meta-Review Compilation.}
The AC integrates insights from Phase I-III discussions, their own observations, and numeric ratings into a meta-review. This document provides a synthesized assessment of the manuscript's strengths and weaknesses that guides the final decision.


\noindent \textbf{V. Paper Decision.}
In the final phase, the AC reviews all meta-reviews for their assigned papers to make an informed decision regarding their acceptance or rejection. We adopt a fixed acceptance rate of 32\%, reflecting the actual average acceptance rate for ICLR $2020\sim2023$. 
Therefore, each AC is tasked with making decisions for a batch of 10 papers and accepts $3\sim4$ papers in the batch.

\subsection{Data Selection} 
\label{sec:data_selection}

The paper data for \model is sourced from real conference submissions to ensure that our simulated reviews closely mirror real scenarios.  
We adhere to four criteria for data selection: 1) The conference must have international impact with a large number of authors and a wide audience, and the academic achievements discussed should have significant real-world impacts; 
2) the papers must be publicly available; 3) the quality of the papers must reflect real-world distribution, including both accepted and rejected papers; 4) the papers must span a broad time range to cover a variety of topics and mitigate the effects of evolving reviewer preferences over time. 

We select ICLR due to its status as a leading publication venue in computer science and its transparency in making both accepted and rejected submissions available. 
We retrieve papers spanning four years (2020$\sim$2023) using OpenReview API\footnote{\url{https://github.com/openreview/openreview-py}}. 
Papers are categorized into oral (top 5\%), spotlight (top 25\%), poster, and rejection. 
We then employ a stratified sampling technique to select papers from each category, resulting in a diverse dataset with 350 rejected papers, 125 posters, 29 spotlights, and 19 orals. This approach ensures the inclusion of papers with varying quality, closely mirroring real-world conferences. 
Finally, we extract the title, abstract, figure and table captions, and the main text that serve as the inputs for the LLM agents. 


\subsection{Baseline Setting}
\label{sec:baseline_setting} 
Real peer review process inherently entails substantial uncertainty due to variations in reviewers' expertise, commitment, and intentions, often leading to seemingly inconsistent numeric ratings. 
For example, NeurIPS experiments found significant differences in reviewers' ratings when different sets of reviewers evaluated the same submissions~\cite{cortes2021inconsistency, zhang2022investigating}. 
Directly comparing numeric ratings of our experimental outcomes with actual ratings can be inappropriate and fail to \emph{disentangle} the latent variables. 

To address this, we establish a \emph{baseline} setting with no specific characteristics of LLM agents (referred to as `\emph{baseline}' in Table~\ref{tab:average_scores}). This allows us to measure the impact of changes in one variable against a consistent reference. 
Across all settings, we generate 10,460 reviews and rebuttals, 23,535 reviewer-AC discussions, 9,414 meta-reviews, and 9,414 paper decisions. Detailed statistics for the dataset are in Appendix Table~\ref{tab:stats}, and the experimental cost is in Appendix~\ref{app:cost}). 

\section{Results}

\subsection{The Role of Reviewers}
\begin{figure}[t]
    \centering
    \includegraphics[width=0.49\linewidth]{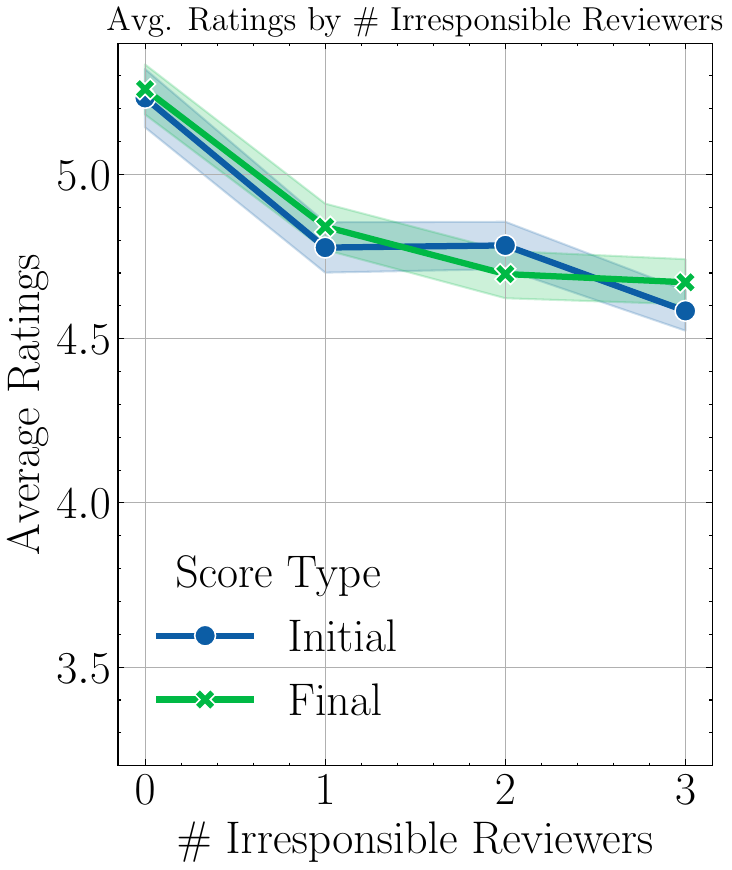}
    \includegraphics[width=0.49\linewidth]{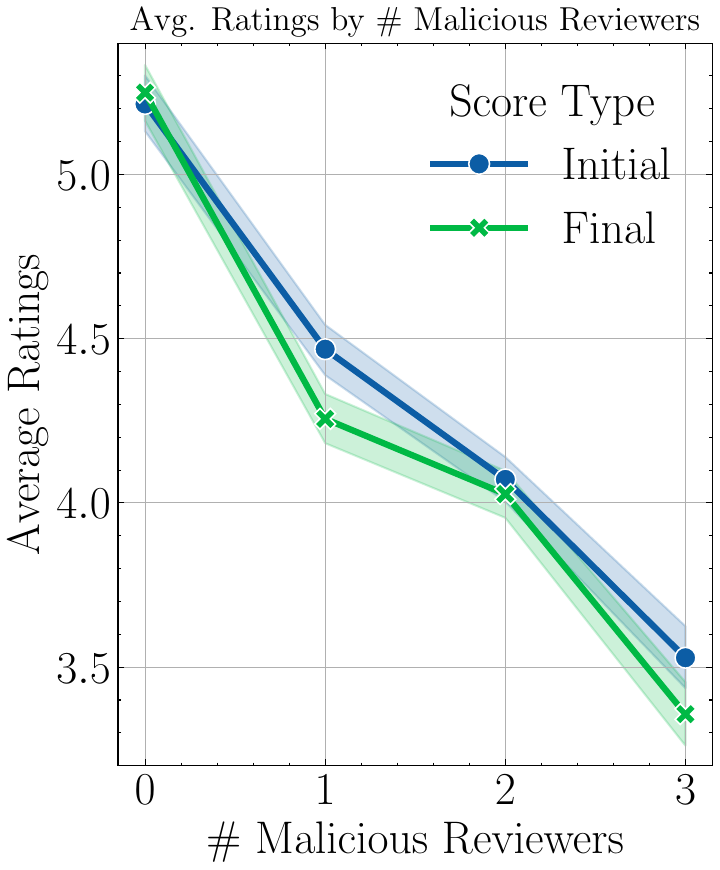}
    \ifthenelse{\boolean{arXivVersion}}
    {
    }
    {\vspace{-3mm}}
    \caption{Distribution of initial and final scores with respect to varying number of irresponsible  \includegraphics[height=1em]{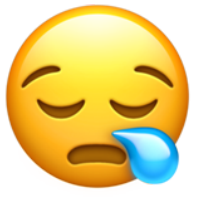} (left) \& malicious \includegraphics[height=1em]{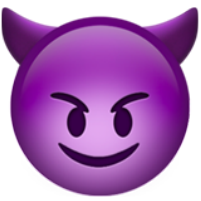} (right) reviewers. 
    }
    \label{fig:vary_num_reviewers}
    \ifthenelse{\boolean{arXivVersion}}
    {
    }
    {
    \ifthenelse{\boolean{arXivVersion}}
    {
    }
    {\vspace{-4mm}}
    }   
\end{figure}
To study the effect of commitment on the peer review outcomes, we start with replacing a \emph{normal} reviewer with either a responsible or an irresponsible reviewer, then gradually increase the number of reviews. 
The settings we consider as well as the initial \& final ratings are in Table~\ref{tab:average_scores}, and the rating distribution is in Figure~\ref{fig:review_ratings}.
Agent-based reviewers in our environment demonstrate classic phenomena in sociology, such as social influence, echo chamber, and halo effects. 

\subsubsection{Overview}
\label{sec:overview}

\begin{table}[!t]
    \centering
    \small
    \begin{tabular}{l|cccc}
    \toprule
    & \multicolumn{2}{c}{Initial (Phase I)} & \multicolumn{2}{c}{Final (Phase III)} \\
    Setting & Avg. & Std. & Avg. & Std. \\ 
        \midrule
        \includegraphics[height=1em]{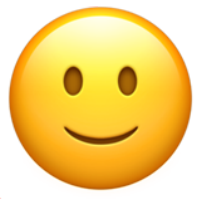} \emph{baseline} & 5.053 & 0.224 & 5.110 & 0.163 \\ 
        \midrule
        \includegraphics[height=1em]{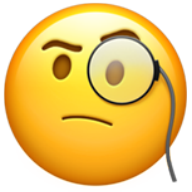} responsible & 4.991 & 0.276 & 5.032 & 0.150 \\ 
        \includegraphics[height=1em]{emoji/irresponsible.pdf} irresponsible & 4.750 & 0.645 & 4.815 & 0.434 \\ 
        \midrule
        \includegraphics[height=1em]{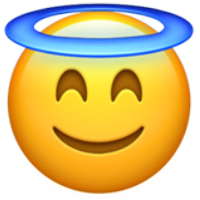} benign & 4.990 & 0.281 & 5.098 & 0.211 \\
        \includegraphics[height=1em]{emoji/malicious.pdf} malicious & 4.421 & 1.181 & 4.368 & 1.014 \\
        \midrule
        \includegraphics[height=1em]{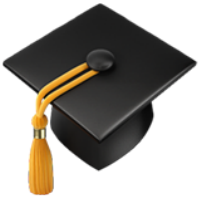} knowledgeable & 5.004 & 0.260 & 5.052 & 0.152 \\ 
        \includegraphics[height=1em]{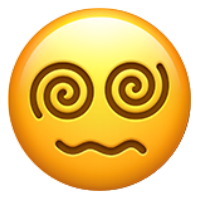} unknowledgeable & 4.849 & 0.479 & 4.987 & 0.220 \\ 
    \bottomrule
    \end{tabular}
    \ifthenelse{\boolean{arXivVersion}}
    {
    }
    {
    \vspace{-.1in}
    }
    \caption{Summary of results. We report the reviewer scores before \& after Reviewer-Author Discussion (Phase III in Figure~\ref{fig:review_pipeline}). `Initial' \& `Final' indicate the reviewer ratings in Phase I \& III, respectively. 
    }
    \label{tab:average_scores}
    \ifthenelse{\boolean{arXivVersion}}
    {
    }
    {
    \vspace{-.2in}
    }   
\end{table}

\noindent \textbf{Social Influence Theory}~\cite{cialdini2004social}
suggests that individuals in a group tend to revise their beliefs towards a common viewpoint. 
A similar tendency towards convergence is also observed among the reviewers. Across all settings, the standard deviation of reviewer ratings (Table~\ref{tab:average_scores}) significant declines after the Reviewer-AC discussion, revealing a trend towards \emph{conformity}. 
This is particularly evident when a highly knowledgeable or responsible reviewer dominates the discussion. 

Overall, responsible, knowledgeable, and benign (well-intentioned) reviewers generally give higher scores than less committed or biased (malicious) reviewers. 
Although initial review ratings can be low, the final ratings in most settings significantly improve following discussions, highlighting the importance of reviewer-author interactions on addressing reviewers' concerns. In Sec.~\ref{sec:peer_review_mechanism}, we further explore whether these interactions and subsequent paper improvements influence the final decisions.  


\subsubsection{Reviewer Commitment}
\label{sec:reviewer_commitment}
\noindent \textbf{Altruism Fatigue \& Peer Effect}~\cite{angrist2014perils} Paper review is typically unpaid and time-consuming~\cite{zhang2021conspicuous}, requiring substantial time investment beyond reviewers' regular professional duties. 
This demanding nature, coupled with \emph{altruism fatigue}---where reviewers feel their voluntary efforts are unrecognized---often results in reduced commitment and superficial assessments. 

The presence of just one irresponsible reviewer can lead to a pronounced decline in overall reviewer commitment compared with the \emph{baseline}. 
Although the initial review length is similar between the two settings (\emph{baseline} and \emph{irresponsible}), averaging 432.4 and 429.2 words, the average word count experienced a significant 18.7\% drop, 
from 195.5 to 159.0 words, after reviewers interact during the reviewer-AC discussion. 
This \emph{peer effect} illustrates how one reviewer's subpar performance can lower the standards and efforts of others, leading to more cursory review post-rebuttal. 
The reduction in overall engagement during critical review discussions underscores the negative impact of insufficient reviewer commitment, which can permit the publication of potentially flawed research, misleading subsequent studies and eroding trust in the academic review process. 

\begin{table}[!t]
    \centering
    \small
    \begin{tabular}{l|lccc}
    \toprule
        Var. & Setting & Jacc. & $\kappa$ & \%Agree \\ 
        \midrule
        \multirow{6}{*}{\includegraphics[height=2em]{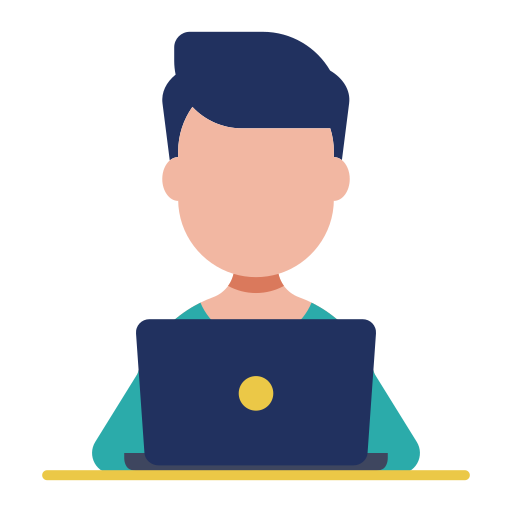}} & \includegraphics[height=1em]{emoji/responsible.pdf} responsible & 0.372 & 0.349 & 72.85 \\ 
        & \includegraphics[height=1em]{emoji/irresponsible.pdf} irresponsible & 0.314 & 0.257 & 69.02 \\ 
        & \includegraphics[height=1em]{emoji/benign.pdf} benign & \textbf{0.632} & \textbf{0.679} & \textbf{86.62} \\ 
        & \includegraphics[height=1em]{emoji/malicious.pdf} malicious & 0.230 & 0.111 & 62.91 \\ 
        & \includegraphics[height=1em]{emoji/knowledgeable.pdf} knowledgeable & 0.297 & 0.230 & 67.88 \\ 
        & \includegraphics[height=1em]{emoji/unknowledgeable.pdf} unknowledgeable & 0.325 & 0.276 & 69.79 \\ 
        \midrule
        \multirow{3}{*}{\includegraphics[height=2em]{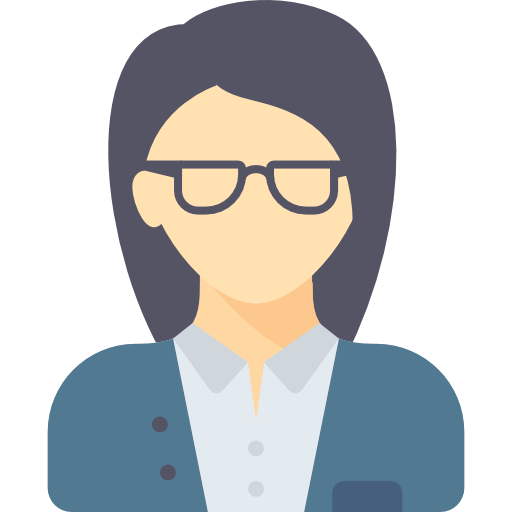}} & \includegraphics[height=1em]{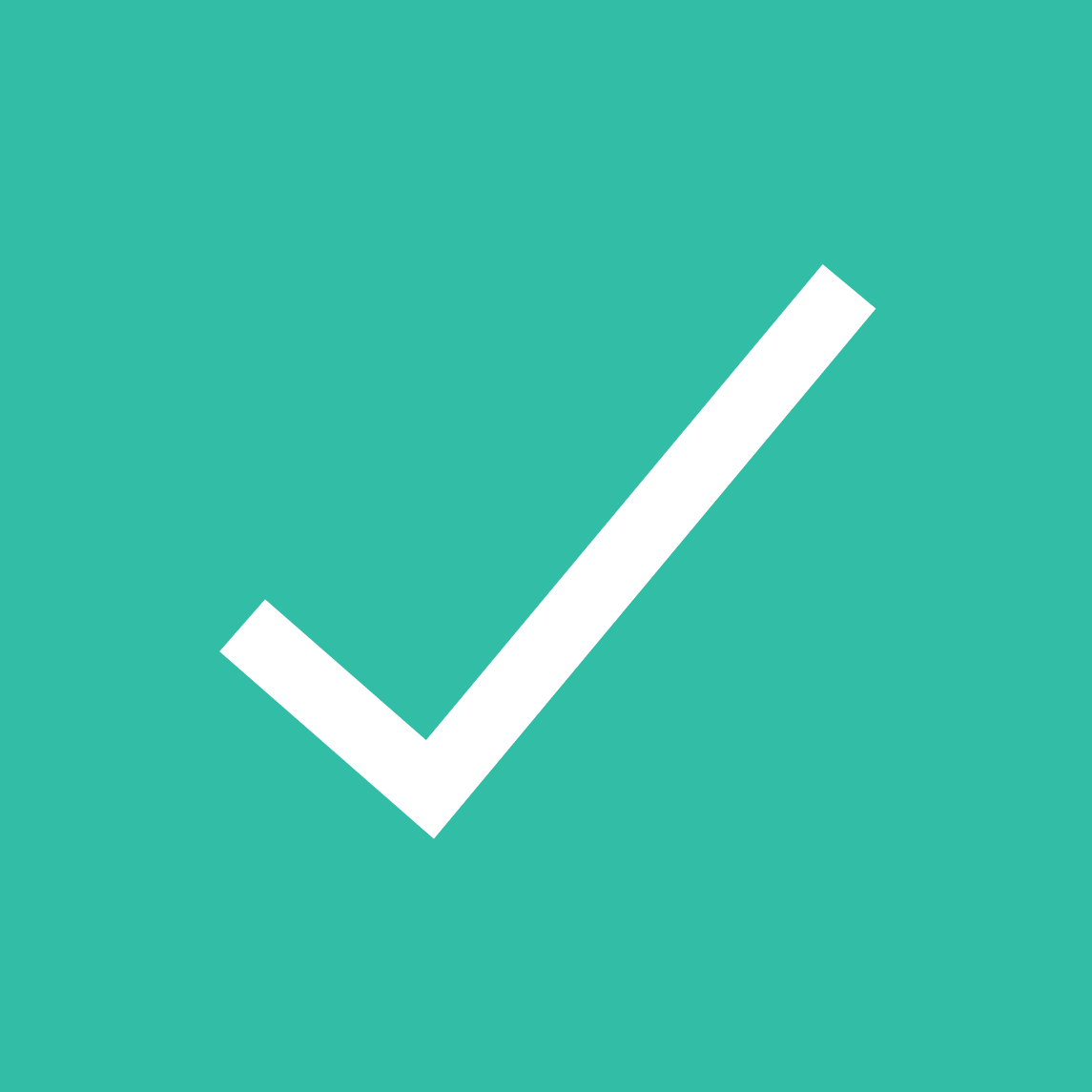} conformist & 0.535 & 0.569 & 82.03 \\ 
        & \includegraphics[height=1em]{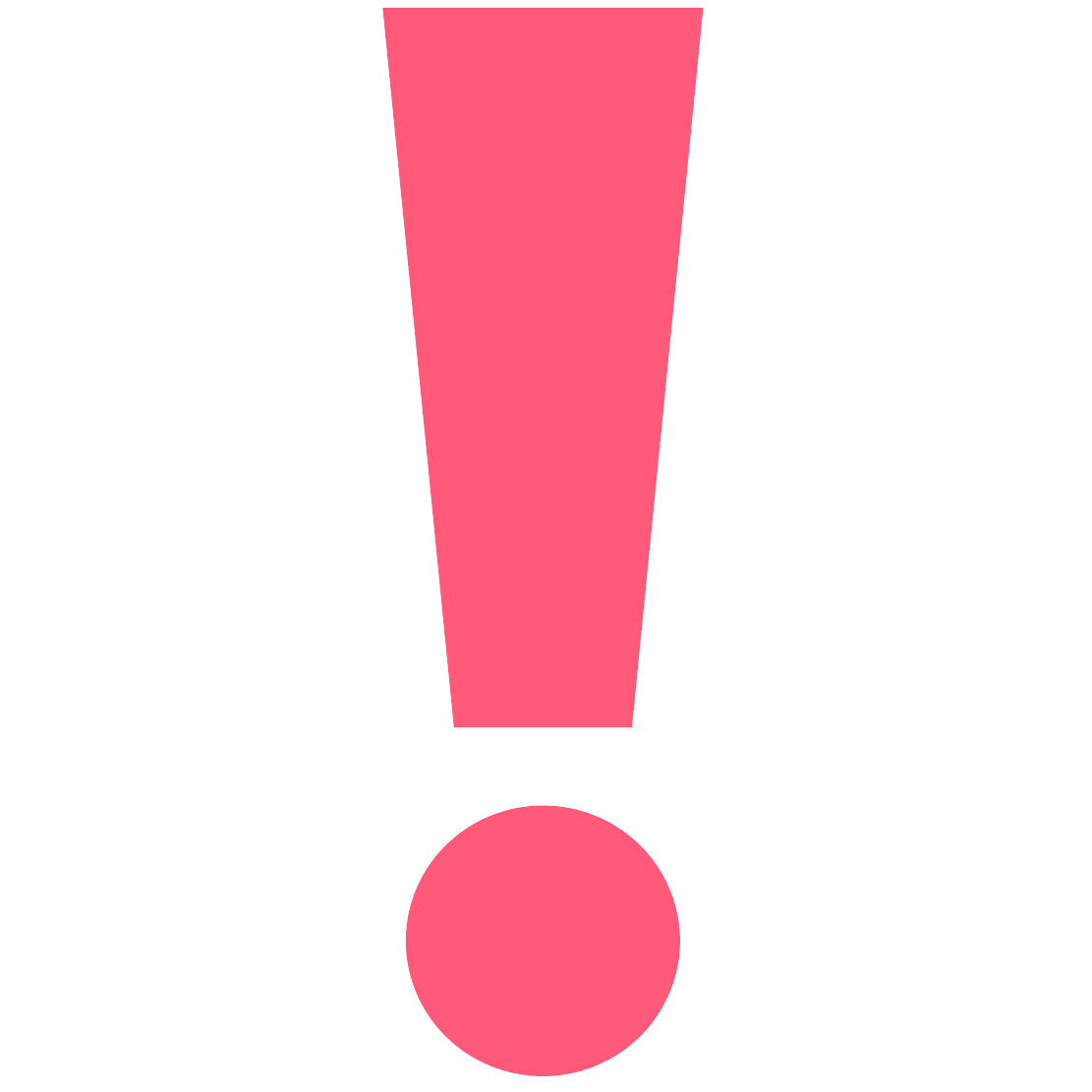} authoritarian & 0.319 & 0.266 & 69.41 \\ 
        & \includegraphics[height=1em]{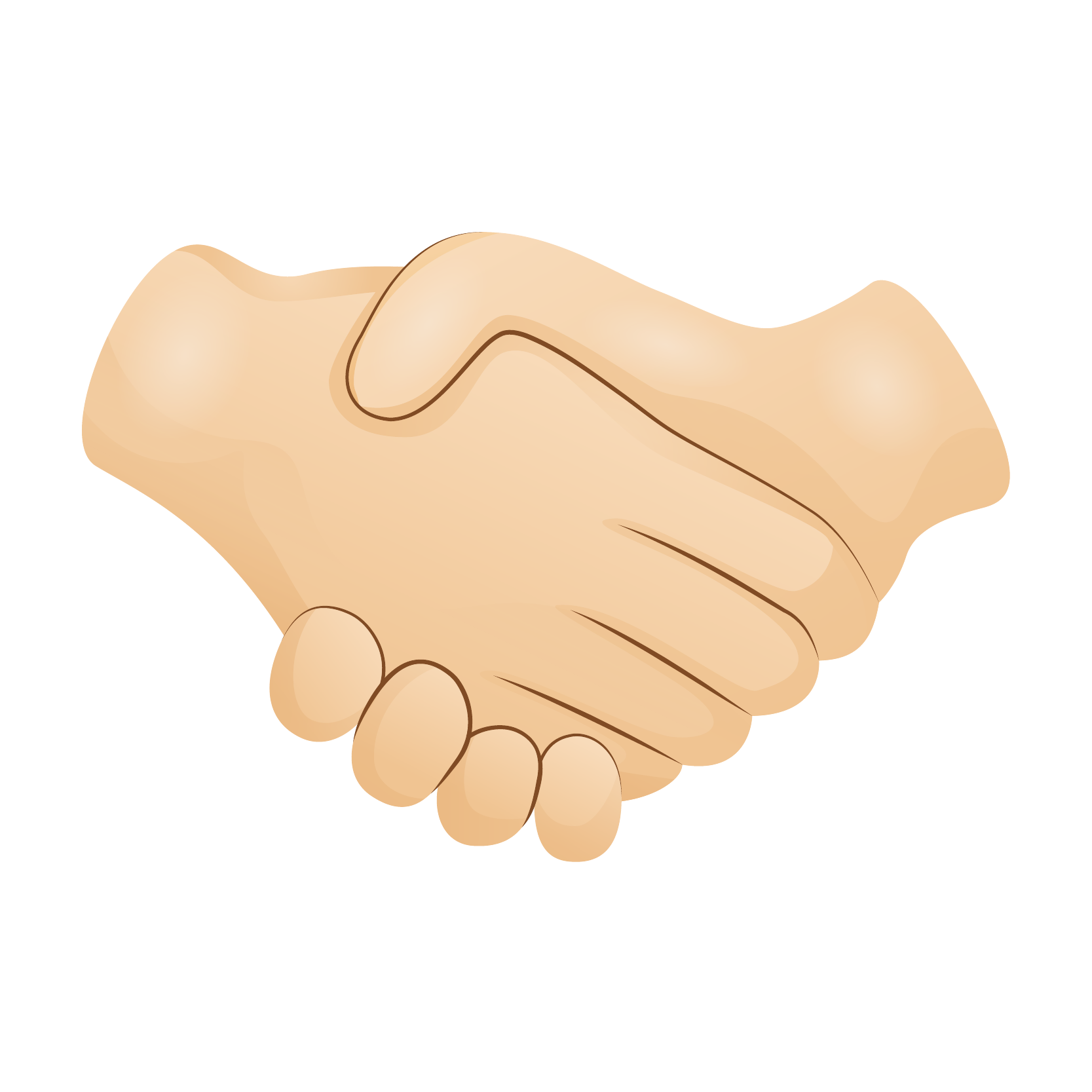} inclusive & 0.542 & 0.578 & 82.41 \\ 
        \midrule
        \multirow{2}{*}{\includegraphics[height=2em]{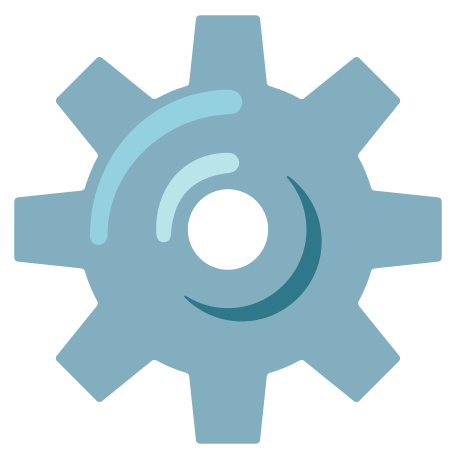}} & \includegraphics[height=1em]{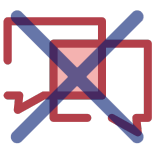} no rebuttals & \underline{0.622} &  \underline{0.668} & \underline{86.14} \\ 
        & \includegraphics[height=1em]{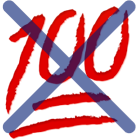} no numeric rating & 0.200 & 0.052 & 60.40 \\ 
        \bottomrule
    \end{tabular}
    \ifthenelse{\boolean{arXivVersion}}
    {
    }
    {\vspace{-.1in}}
    \caption{Comparison of final decisions in various settings relative to the \emph{baseline} experiment in terms of Jaccard Index (Jacc.), Cohen's Kappa Coefficient ($\kappa$), and Percentage Agreement (\%Agree). Jacc. indicate the set of papers accepted by both the investigated setting and the baseline. The highest and second highest values are highlighted in \textbf{bold} and \underline{underlined}, respectively.
    }
    \label{tab:agreement_with_baseline}
    \ifthenelse{\boolean{arXivVersion}}
    {
    }
    {\vspace{-.25in}}
\end{table}

\noindent \textbf{Groupthink}~\cite{janis2008groupthink}
occurs when a group of reviewers, driven by a desire for harmony or conformity, reaches a consensus without critical reasoning or evaluation of a manuscript. It can be especially detrimental when the group includes irresponsible or malicious reviewers. To examine such effects, we substitute $1\sim3$ normal reviewers with irresponsible reviewers and analyze the changes in ratings before \& after reviewer-AC discussion. 

Table~\ref{tab:score_change_irresponsible} highlights a noticeable decline in review ratings under the influence of irresponsible reviewers. Replacing 2 normal reviewers with irresponsible ones results in a significant drop of 0.25 from 5.256 to 5.005 in the average reviewer rating after Reviewer-AC Discussion (Phase III). In contrast, in the \emph{baseline} scenario, the final ratings improve by an average 0.06 post-rebuttal, as reviewers more proactively scrutinize the author feedback and have their concerns addressed. Interestingly, the scores among irresponsible reviewers exhibit a slight increase, suggesting a tendency to conform to the assessments of normal reviewers. 

\begin{figure*}[t]
    \centering
    \includegraphics[width=0.95\linewidth]{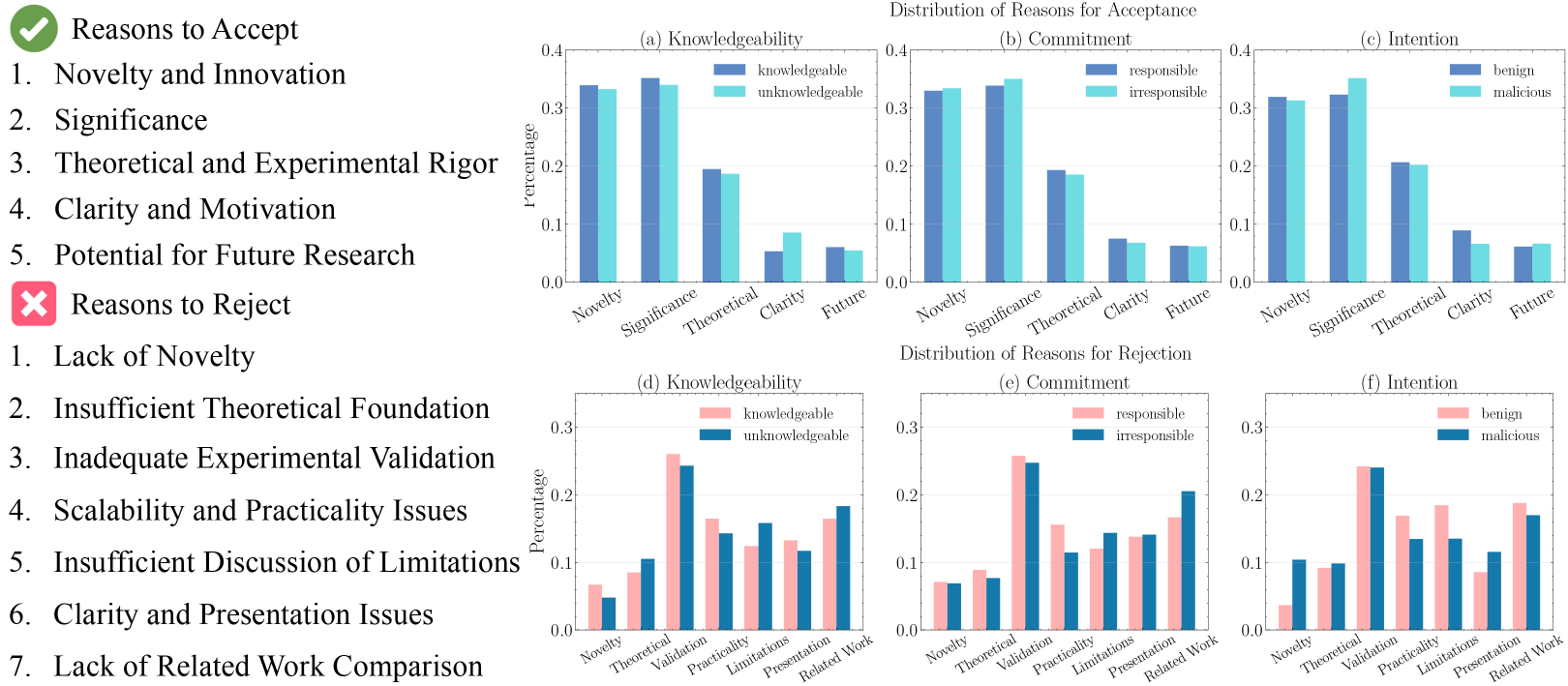}
    \ifthenelse{\boolean{arXivVersion}}
    {
    }
    {\vspace{-2mm}}
    \caption{Distribution of reasons for acceptance and rejections. }
    \label{fig:accept_reject_reasons_distribution}
    \ifthenelse{\boolean{arXivVersion}}
    {
    }
    {
    \vspace{-2mm}
    }
\end{figure*}



\subsubsection{Reviewer Intention}
\label{sec:reviewer_intention}
\noindent \textbf{Conflict Theory}~\cite{bartos2002using} states that 
societal interactions are often driven by conflict rather than consensus. 
In the context of peer review, where the acceptance of papers is competitive, 
reviewers may perceive other high-quality submissions as threats to their own work due to conflict of interests. 
This competitive behavior can lead to low ratings for competing papers, particularly for concurrent works with highly similar ideas, as reviewers aim to protect their own standing in the field. 
Empirically, the reviewer ratings in Figure~\ref{fig:review_ratings} show a significant shift to a bimodal distribution, primarily centered around $[4.0, 4.25]$, when just one 
malicious reviewer is involved. This forms a stark contrast to the unimodal distribution between $[5.0, 5.25]$ observed in the \emph{baseline} condition. 

\noindent \textbf{Echo Chamber Effects}~\cite{cinelli2021echo} occur when  a group of reviewers sharing similar biases amplify their opinions, leaning towards a collective decision without critically evaluating merits of the work. 
As illustrated in Figure~\ref{fig:vary_num_reviewers}, increasing the number of malicious reviewers from 0 to 3 results in a consistent drop in the average rating from 5.11 to 3.35, suggesting that the presence of malicious reviewers significantly impacts the overall evaluation. 
Meanwhile, as malicious reviewers predominate, the average rating among these biased reviewers (Table \ref{tab:score_change_malicious}) experiences a greater drop post-rebuttal, indicating that the inclusion of more biased reviewers not only amplifies the paper's issues but also solidifies their strong negative opinions about the work. 
This process not only reinforces pre-existing biases and reduces critical scrutiny, but also has a spillover effect that adversely impacts evaluations from unbiased reviewers. The presence of 1 and 2 \emph{malicious} reviewers corresponds to a decline by 0.14 and 0.10, respective, among the normal reviewers.  


\noindent \textbf{Content-level Analysis} We categorize the reasons for acceptance and rejection as shown in Figure~\ref{fig:accept_reject_reasons_distribution} with additional details provided in Appendix~\ref{app:summarize}. 
While reasons for accepting the papers are consistent across all settings, the reasons for rejection differ significantly in distribution. 
Irresponsible reviews tend to be shallow,  cursory, and notably 22.2\% shorter, whereas malicious reviews disproportionally criticize the \emph{lack of novelty} in the work (Figure~\ref{fig:accept_reject_reasons_distribution}d), a common but vague reason for rejection. Specifically, mentions of \emph{lack of novelty} by \emph{malicious} reviewers account for 10.4\% of feedback, marking a 182.9\% increase compared to just 3.69\% by \emph{benign} reviewers. 
They also highlight more \emph{presentation} issues which, although important for clarity, do not pertain to the theoretical soundness of the research. 
On the other hand, benign reviewers tend to focus more on discussions about \emph{scalability and practicality} issues, providing suggestions to help enhance papers' comprehensiveness.

\begin{table*}[!ht]
    \centering
    \resizebox{.7\textwidth}{!}{
    \begin{tabular}{lllc|lllc}
        \toprule
        \multicolumn{4}{c}{\includegraphics[height=1em]{emoji/normal.pdf} normal reviewers} & \multicolumn{4}{|c}{\includegraphics[height=1em]{emoji/irresponsible.pdf} irresponsible reviewers} \\ 
        \# & Initial & Final & $+/-$ & \# & Initial & Final & $+/-$ \\ 
        \midrule
        3 & 5.053 $\pm$ 0.623 & 5.110 $\pm$ 0.555 & $+$0.06 & 0 & / & / & / \\
        2 & 5.056 $\pm$ 0.633 & 5.015 $\pm$ 0.546 & $-$0.04 & 1 & 4.139 $\pm$ 1.121 & 4.416 $\pm$ 0.925 & $+$0.27 \\
        1 & 5.256 $\pm$ 0.896 & 5.005 $\pm$ 0.630 & $-$0.25 & 2 & 4.548 $\pm$ 0.925 & 4.543 $\pm$ 0.872 & $-$0.01 \\
        0 & / & / & / & 3 & 4.591 $\pm$ 0.912 & 4.677 $\pm$ 0.745 & $+$0.09 \\
        \bottomrule
        \end{tabular}
        }
        \ifthenelse{\boolean{arXivVersion}}
        {
        }
        {
        \vspace{-2mm}
        }
        \caption{Average reviewer ratings when varying numbers of \includegraphics[height=1em]{emoji/normal.pdf} \emph{normal} reviewers are replaced by \includegraphics[height=1em]{emoji/irresponsible.pdf} irresponsible reviewers. `\#' represents the number of reviewers of each type. `Initial' \& `Final' refer to the average ratings in Phase I \& III. The left and right side of the table shows average ratings from \includegraphics[height=1em]{emoji/normal.pdf} normal reviewers and \includegraphics[height=1em]{emoji/irresponsible.pdf} irresponsible reviewers, respectively. $+/-$ indicates the change in average ratings after rebuttals. }
    \label{tab:score_change_irresponsible}
\vspace{-0.12in}
\end{table*}

\subsubsection{Reviewer Knowledgeability} 


Knowledgeability poses two challenges.  
Firstly, despite extended efforts at matching expertise, review assignments are often imperfect or random \cite{xu2024one, saveski2024counterfactual}. Secondly, the recent surge in submissions to computer science conferences has necessitated an expansion of the reviewer pools, raising concerns about the adequacy of reviewers' expertise to conduct proper and effective evaluations. 
As shown in Figure~\ref{fig:accept_reject_reasons_distribution}, less knowledgeable reviewers are 24\% more likely to mention \emph{insufficient discussion of limitations}, whereas expert reviewers not only address these basic aspects but also provide 6.8 \% more critiques on experimental validation, resulting in more concrete and beneficial feedback for improving the paper. 

\subsection{Involvements of Area Chairs}

\begin{figure}[t]
    \centering
    \includegraphics[width=0.96\linewidth]{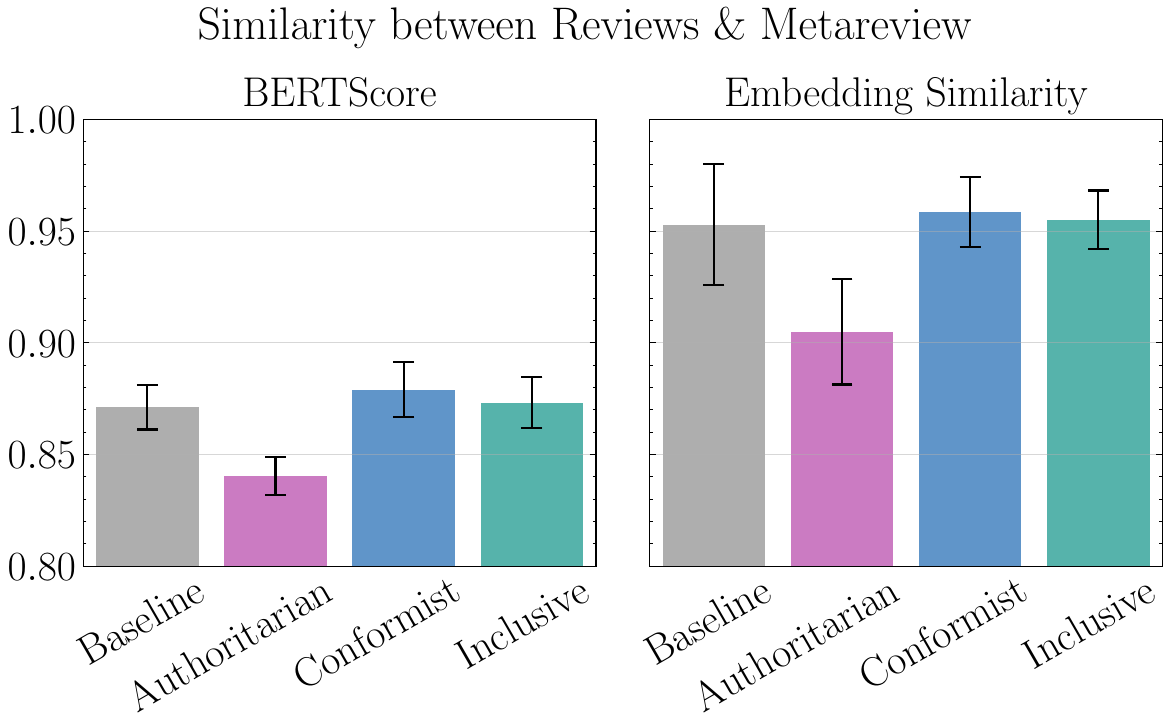} \\
    \vspace{-3mm}
    \caption{Similarities between reviews and meta-reviews w/ various intervention strategies from AC. Left: BERTScore, right: sentence embedding similarity.}
    \label{fig:similarity_review_metareview}
    \vspace{-4mm}
\end{figure}

We quantify the alignment between reviews and meta-reviews using BERTScore~\cite{zhangbertscore} and sentence embedding similarity~\cite{reimers2019sentence} in Table~\ref{tab:agreement_with_baseline}, and measure the agreement of final decisions between \emph{baseline} and each setting in Figure~\ref{fig:similarity_review_metareview}. 
Inclusive ACs align most closely with the \emph{baseline} for final decisions, 
demonstrating their effectiveness in integrating diverse viewpoints and maintaining the integrity of the review process through a balanced consideration of reviews and their own expertise. 
In contrast, authoritarian ACs manifest significantly lower correlation with the \emph{baseline}, with a Cohen's Kappa of only 0.266 and an agreement rate of 69.8\%. This suggests that their decisions may be skewed by personal biases, leading to acceptance of lower quality papers or the rejection of high-quality papers that do not align with their viewpoints, thereby  compromising the integrity and fairness of the peer review process. 
Conformist ACs, while showing a high semantic overlap with reviewers' evaluations as evidenced in Figure~\ref{fig:similarity_review_metareview}, might lack independent judgment. This dependency could perpetuate existing biases or errors in initial reviews, underscoring a critical flaw in overly deferential approaches. 

\subsection{Impacts of Author Anonymity}
\label{sec:author_anonymity}
Recent conferences have increasingly permitted the release of preprints, potentially impacting paper acceptance~\cite{elazar2024estimating}.  Although reviewers are instructed not to proactively seek information about author identities, concerns persist that reviews may still be biased by author reputation. 

\noindent \textbf{Authority bias} is the tendency to attribute greater accuracy and credibility to the opinions of authority figures. 
This bias is closely related to the \textbf{Halo Effects}, a cognitive bias where the positive perception of an individual in one area, such as their previous groundbreaking research, influences judgments about their current work. 
Reviewers influenced by authority bias are more likely to give favorable reviews to well-known and respected scientists. 
 
To analyze the impact of author identities on review outcomes, we vary the number of reviewers aware of the authors' identities ($k$), ranging from 1 to 3, and adjusted the proportion of papers with known author identities ($r$) from 10\% to 30\%. 
Specifically, the reviewers were informed that the authors of certain papers were renowned and highly accomplished in the field.  
We categorized papers into two types: higher quality and lower quality, based on their ground-truth acceptance decisions. 

\begin{figure}[t]
    \centering
    \includegraphics[width=0.96\linewidth]{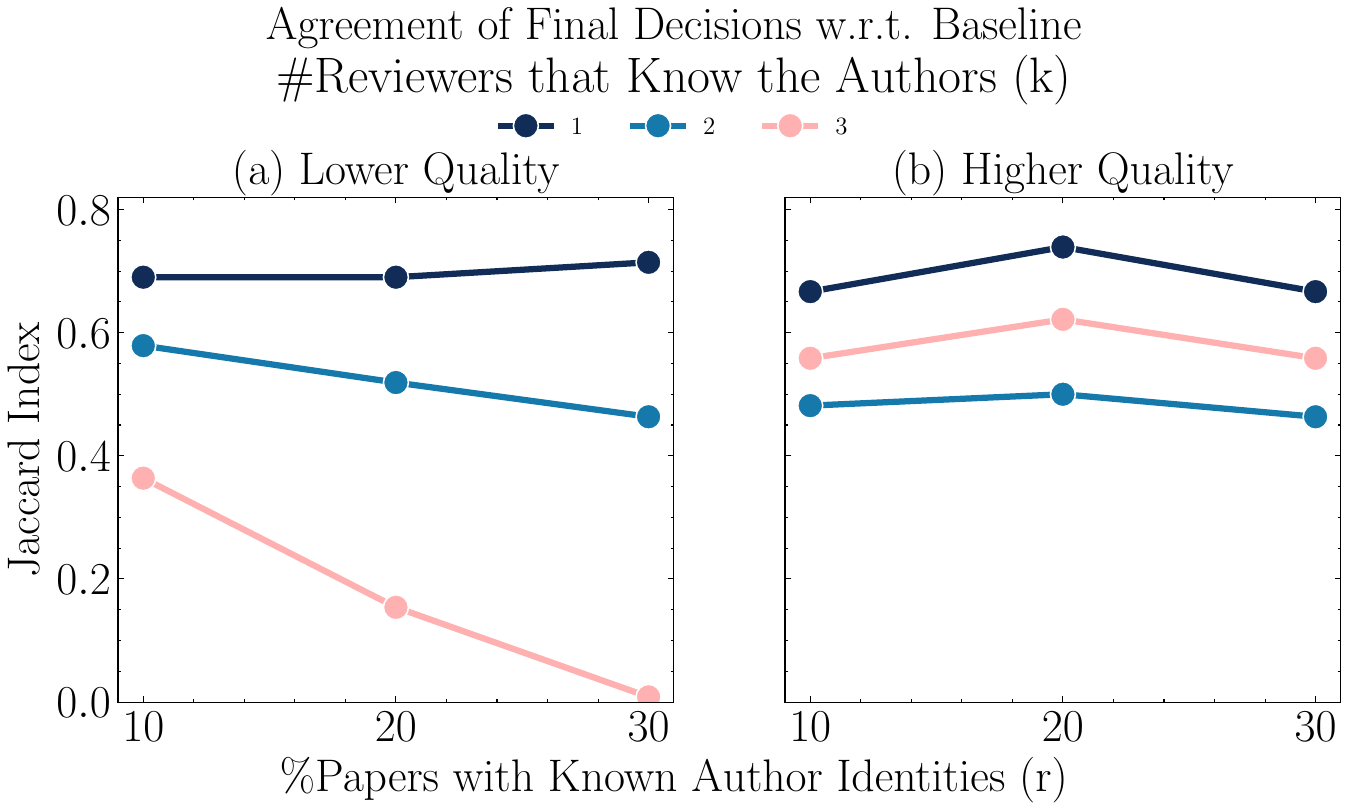}
    \ifthenelse{\boolean{arXivVersion}}
    {
    }
    {
    \vspace{-3mm}
    }
    \caption{Comparison of final decisions with respect to \emph{baseline} when the author identity is known for varying ratios of papers, relative to the \emph{baseline}. 
    Smaller Jaccard indices suggest lower correlation with the baseline. }
    \label{fig:known_authors_jacc}
    \ifthenelse{\boolean{arXivVersion}}
    {
    }
    {
    \vspace{-2mm}
    }
\end{figure}

For lower-quality papers, awareness of the authors' renowned identities among 1, 2, or 3 reviewers resulted in Jaccard indices of 0.364, 0.154, and 0.008, respectively, in terms of paper acceptance (Figure~\ref{fig:known_authors_jacc}). The most extreme case has a negative Cohen's Kappa $\kappa$ (Figure~\ref{fig:known_authors_kappa}), indicating a substantial deviation in paper decisions. When high-quality papers had known author identities, much less significant changes were observed in accepted papers. Notably, changes in paper decisions are more influenced by the number of reviewers aware of the author identities than by the percentage of papers with known author identities. 

\subsection{Effects of Peer Review Mechanisms}
\label{sec:peer_review_mechanism}
We investigate two variations to peer review mechanisms. 1) \emph{no rebuttal}---excluding the Reviewer-Author Discussion (Phase II) and the Reviewer-AC Discussion (Phase III); 2) \emph{no numeric rating}---removing the requirement to assign overall numeric ratings (Phase I \& III), thus making the AC's decision solely dependent on the content of the reviews.

\noindent \textbf{Effects of Rebuttals.}
Eliminating the rebuttal phase, which requires substantial time commitments from both reviewers and authors, has a surprisingly minimal impact on the final paper decisions, aligning closely with the \emph{baseline} scenario. 

One explanation for this minimal impact is the \emph{anchoring bias}, where the initial impression formed during the first submission (the ``anchor'') predominantly influences reviewers' judgments. 
Even though authors may make substantial improvements during the rebuttal phase that address reviewers' concerns  (Sec.~\ref{sec:overview}), these changes may fail to alter their initial judgments. Another plausible reason is that all submissions improve in quality during the rebuttal phase. Thus, the relative position (ranking of quality) of each paper among all submissions experiences little change. 

\noindent \textbf{Effects of Overall Ratings.} 
Numeric ratings from reviewers may serve as a shortcut in the final decision-making process for paper acceptance. 
When these ratings are omitted, the decision-making landscape changes significantly, leading to potentially divergent decisions. 
The comparison of outcomes with respect to \emph{baseline} reveals only a minimal overlap, with a Jaccard index of 0.20 in terms of accepted papers (Table~\ref{tab:agreement_with_baseline}). 

\section{Related Work}
\vspace{-2mm}
\textbf{Analysis of Peer Review Systems.}
Peer review serves as the backbone of academic research, ensuring the integrity and quality of published work~\cite{zhang2022system}. 
Several studies have scrutinized various challenges within peer review, such as bias~\cite{stelmakh2021prior, ugarov2023peer, verharen2023chatgpt, liu2023testing}, conflict of interests~\cite{mcintosh2023safeguarding}, and the broader issues of review quality and fairness~\cite{stelmakh2021prior, mcintosh2023safeguarding, stephen2024distinguishing}.  
Research has also delved into the operational aspects, such as reviewer assignments~\cite{jovanovic2023reviewer, saveski2024counterfactual, kousha2024artificial} and author rebuttals~\cite{huang2023makes}, identifying areas for improvement in transparency, fairness, and accountability~\cite{zhang2022investigating}. These studies primarily focus on analyzing existing real-world review data and outcomes. 
However, due to the complexity and inherent variability of peer review, isolating the effects of specific factors on review outcomes remains a significant challenge. 

\noindent \textbf{LLMs as Agents.}
Large Language Models (\llms) such as GPT-4~\cite{openai2023gpt4}, Claude 3~\cite{claude3}, and Gemini~\cite{team2023gemini} have not only demonstrated sophisticated language understanding and generation skills~\cite{xiong2024search}, but also exhibit planning, collaboration, and competitive behaviors~\cite{zhao2023competeai, bai2023benchmarking}. 
Our study aligns with recent works in agent-based modeling (ABM), such as ChatArena~\cite{ChatArena}, ChatEval~\cite{chan2023chateval}, Lumos~\cite{yin2023lumos}, and MPA~\cite{zhu2024dynamic}, that leverage the capabilities of LLM agents to simulate realistic environments for scientific research~\cite{li2023prd,li2024embodied,jiang2024multi,chan2023chateval,xie2024can}. 

\vspace{-2mm}
\section{Conclusion}
\vspace{-2mm}
We presented \model, the first LLM-based framework for simulating the peer review process. 
\model addresses key challenges by disentangling intertwined factors that impact review outcomes while preserving reviewer privacy.
Our work lays a solid foundation for more equitable and transparent review mechanism designs in academic publishing. 
Future works could investigate how intricate interactions between different variables collectively affect review outcomes. 

\section*{Limitation}

Our work has the following limitations. First, \model is unable to dynamically incorporate or adjust experimental results in response to reviewer comments during Reviewer-Author Discussion (Phase II in Figure~\ref{fig:review_pipeline}), as LLMs lack the capability to generate new empirical data. 
Secondly, our analysis mainly isolates and examines individual variables of the peer review process, such as reviewer commitment or knowledgeability. Real-world peer reviews, however, involve multiple interacting dimensions. 
Finally, we did not directly compare the simulation outcomes with actual peer review results. 
As described in Sec~\ref{sec:baseline_setting}, 
establishing a consistent baseline for such comparisons is challenging due to the wide variability in human reviewer characteristics, such as commitment, intention, and knowledgeability, which can vary across papers, topics, and time periods. 
The inherent variability and arbitrariness in human peer reviews~\cite{cortes2021inconsistency} add complexity to direct comparisons between simulated and real outcomes.



\section*{Ethical Consideration}

\noindent \textbf{Further Investigation into Peer Review data.}
The sensitivity and scarcity of real-world review data complicate comprehensive studies of peer reviews due to ethical and confidentiality constraints. Our \model framework generates simulated data to study various peer review dynamics, effectively overcoming related challenges. 

\noindent \textbf{Peer Review Integrity.} 
As discussed, the integrity of the peer review process is underpinned by the commitment, intention, and knowledgeability of reviewers. \emph{Knowledgeability} ensures that reviewers can accurately assess the novelty,  significance, and technical soundness of submissions. 
Good \emph{intention} are essential for maintaining the objectivity and fairness of reviews, thereby supporting the credibility and integrity of academic publications. 
A high level of \emph{commitment} from reviewers ensures comprehensive and considerate evaluations of submission, which is important for a fair and rigorous evaluation process. However, paper review is usually an unpaid and time-consuming task. Such demanding nature can lead the reviewers to conduct cursory or superficial evaluations. 

\noindent \textbf{Caution about Use of LLMs.} 
Our \model mirrors real-world academic review practices to ensure the authenticity and relevance of our findings. 
While \model uses LLMs to generate paper reviews, there are ethical concerns regarding their use in actual peer review processes~\cite{lee2023surveying}. Recent machine learning conferences have shown an increase in reviews suspected to be AI-generated~\cite{liang2024monitoring}. 
Although LLM-generated reviews can provide valuable feedback, we strongly advise against their use as replacements for human reviewers in real-world peer review processes. As LLMs are still imperfect, human oversight is crucial for ensuring fair and valuable assessments of manuscripts and for maintaining the integrity and quality of peer reviews. 



\clearpage

\bibliography{custom}

\appendix

\onecolumn
\addcontentsline{toc}{section}{Appendix} 
\part{Appendix} 
\parttoc 

\section{Experimental Details}
\label{sec:appendix}

\subsection{Review Categorization}
\label{app:summarize}

In our experiment, we utilize GPT-4 to summarize and categorize the reasons for paper acceptance and rejection, as illustrated in Figure~\ref{fig:accept_reject_reasons_distribution}. Specifically, we analyze each line from the \emph{reasons for acceptance} and \emph{reasons for rejection} fields in the generated reviews. GPT-4 is tasked with automatically classifying each listed reason. If an entry does not align with predefined categories, the model  establish a new category. Ultimately, we identify five distinct reasons for acceptance and seven reasons for rejection.

\begin{table}[htbp]
    \centering
    \begin{tabular}{lcc}
    \toprule
        ~ & \#Words & \#Characters \\ 
        \midrule
        Review & 438.2 $\pm$ 72.0 & 3067.4 $\pm$ 510.1 \\ 
        Rebuttal & 370.6 $\pm$ 49.9 & 2584.8 $\pm$ 376.5 \\ 
        Updated Review & 189.7 $\pm$ 46.6 & 1304.0 $\pm$ 320.8 \\ 
        Meta-review & 256.9 $\pm$ 64.8 & 1849.9 $\pm$ 454.5 \\ 
    \bottomrule
    \end{tabular}
    \caption{Statistics of our dataset.}
    \label{tab:stats}
\end{table}


\begin{figure}[h]
    \centering
    \includegraphics[width=0.5\linewidth]{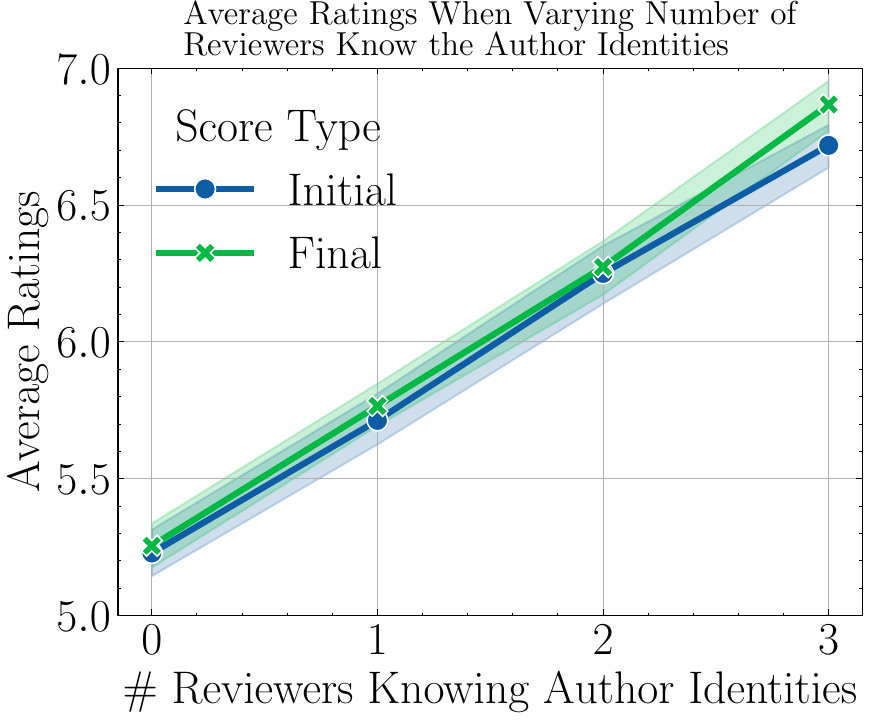}
    \ifthenelse{\boolean{arXivVersion}}
    {
    }
    {\vspace{-3mm}}
    \caption{Distribution of initial and final ratings when varying numbers of reviewers are aware of the authors' prestigious identity.  
    }
    \label{fig:vary_num_reviewers_knowing_author_identities}
    \ifthenelse{\boolean{arXivVersion}}
    {
    }
    {\vspace{-4mm}}
\end{figure}

\subsection{Experimental Costs} 
\label{app:cost}
To ensure consistent evaluation results, we use the \textsf{gpt-4-1106-preview} version of the GPT-4 model throughout our experiments. The model is selected for its superior language understanding and generation capabilities, essential for simulating an authentic peer review process. 
To enhance reproducibility and minimize API usage, we establish a \emph{baseline} settings (Sec.~\ref{sec:baseline_setting}), where no specific personalities of the role are detailed (`\emph{baseline}' in Table~\ref{tab:average_scores}). This setting allows us to measure the impact of changes in individual variables against a consistent standard. For subsequent experiments, we adopt reviews and rebuttals (Phase I-II) from this \emph{baseline} when applicable. 
For example, when we investigate the effects of substituting a normal reviewer with an irresponsible person, we only generate the reviews for that specific reviewer while adopting existing reviews from the \emph{baseline} setting. This approach minimizes the variability caused by different experimental runs and significantly reduces the API cost compared with rerunning the entire review pipeline each time. 
The total cost of API usage across all tests is approximately \$2780. 


\subsection{Model Selection} 
\label{app:model_selection}
Additionally, we have also explored the feasibility of alternative models, such as \textsf{gpt-35-turbo} and Gemini. These models were initially considered to assess the cost-effectiveness and performance diversity. However, these models either encounter issues related to content filtering limitations, resulting in the omission of critical feedback, or generate superficial evaluations and exhibited a bias towards overly generous scoring. Therefore, despite the higher operational costs, we choose despite the higher operational costs, due to its more consistent and realistic output in peer review simulations due to its more consistent and realistic output in peer review simulations.

\subsection{Behavioral Analysis of LLM Agents} 
\label{app:behavioral_analysis}

\paragraph{Qualitative Evidence} 
Table~\ref{tab:review_example} presents the LLM-generated review, rebuttal, and meta-review for the paper \texttt{Image as Set of Points}~\cite{ma2022image}, demonstrating substantial overlap with human reviews in Table~\ref{tab:real_review_example}. 

\paragraph{Quantitative Evidence}
We randomly sample 100 papers from our dataset, use LlamaIndex~\footnote{\url{https://www.llamaindex.ai/}} to extract and match major comments in human and LLM-generated reviews in our dataset. To ensure fairness, we follow~\citeauthor{liang2023can} and ask the LLM reviewers to generate 4 reasons to accept / reject for each paper. 
In 90\% / 77\% / 39\% of the papers, at least 2 / 3 / 4 out of 4 points align with human reviewers, indicating that LLMs provide realistic opinions. Moreover, LLMs highlight unique insights often overlooked by human reviewers, such as computational costs, scalability concerns, and experiments on diverse datasets.

\subsection{Additional Results and Statistics}

\begin{figure}[]
    \centering
    \includegraphics[width=0.8\linewidth]{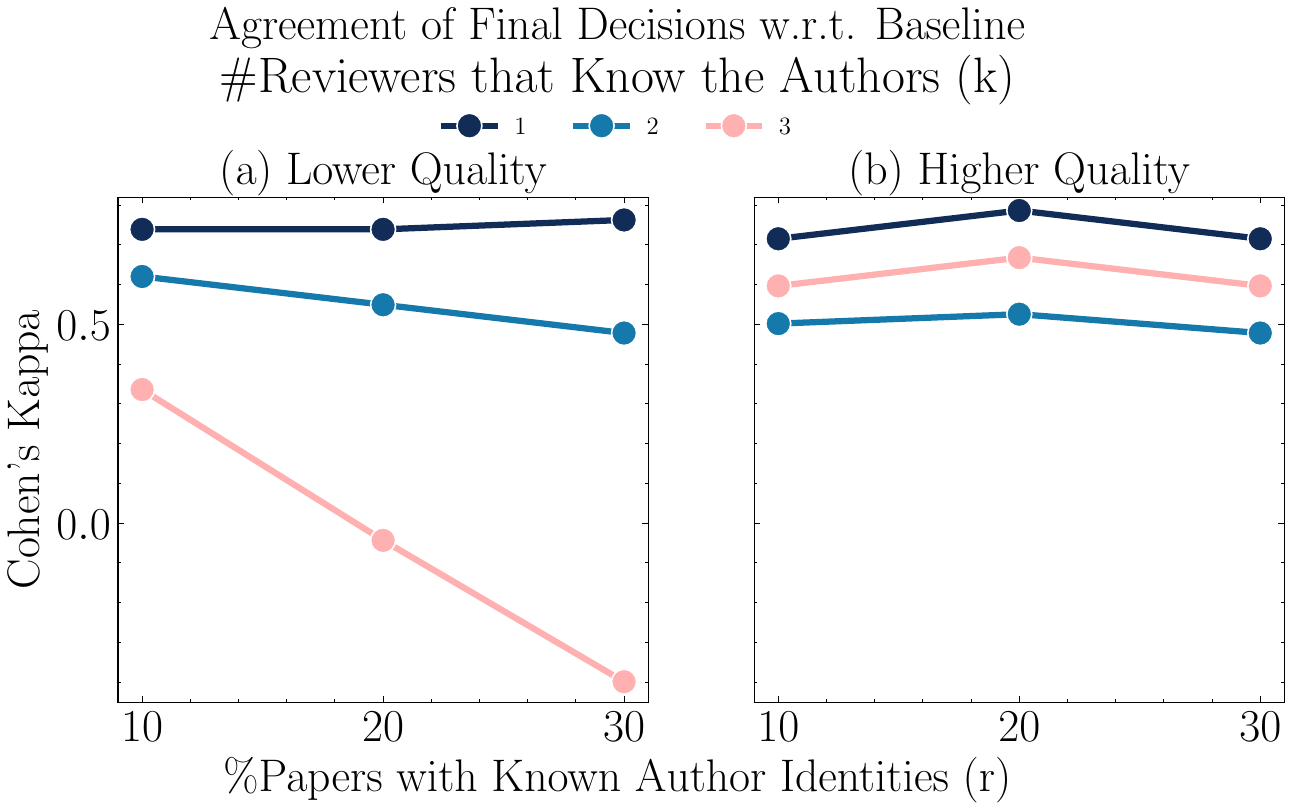}
    \ifthenelse{\boolean{arXivVersion}}
    {
    }
    {\vspace{-3mm}}
    \caption{Comparison of final decisions with respect to \emph{baseline} when the author identity is known for varying ratios of papers, relative to the \emph{baseline}. 
    A smaller Cohen's Kappa coefficient suggests a lower correlation with the baseline. }
    \label{fig:known_authors_kappa}
    \ifthenelse{\boolean{arXivVersion}}
    {
    }
    {\vspace{-.2in}}
\end{figure}

\begin{figure}[htbp]
    \centering
    \includegraphics[width=0.98\linewidth]{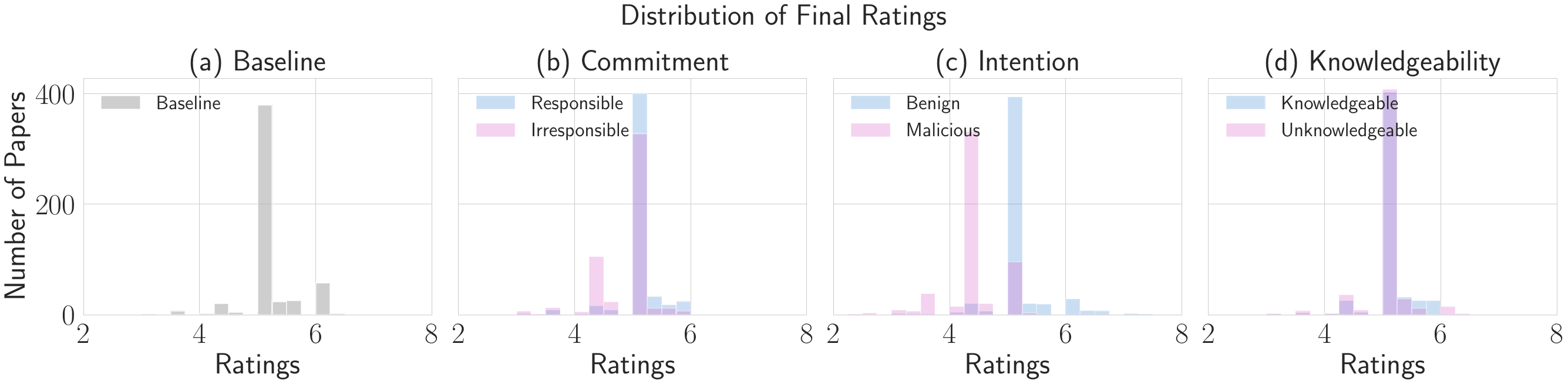}
    \ifthenelse{\boolean{arXivVersion}}
    {
    }
    {\vspace{-3mm}}
    \caption{Final rating distribution when we vary one reviewer in the experiment, including their commitment, intention, and knowledgeability.}
    \label{fig:review_ratings}
    \ifthenelse{\boolean{arXivVersion}}
    {
    }
    {\vspace{-4mm}}
\end{figure}

\begin{itemize}[leftmargin=1em]
    \setlength\itemsep{0em}
    \item Table~\ref{tab:stats} is the statistics of our dataset, including the word and character counts of the generated reviews, rebuttals, updated reviews, and meta-reviews. 
    \item Table~\ref{tab:score_change_malicious} is the average reviewer ratings when varying number of \emph{normal} reviewers are replaced by \emph{malicious} reviewers.
    \item Table~\ref{fig:review_ratings} shows the prompts used in \model and the characteristics of each type of roles. 
    \item Figure~\ref{fig:vary_num_reviewers_knowing_author_identities} is the distribution of initial and final ratings as $0 \sim 3$ reviewers become aware of the authors' prestigious identity. It shows that the average reviewer ratings consistently increase with more reviewers knowing the author identities. Meanwhile, reviewer ratings consistently increase after rebuttals.
    \item Figure~\ref{fig:known_authors_kappa} is the Cohen's Kappa coefficient ($\kappa$) when the author identity is known for varying ratios of papers, relative to the \emph{baseline}. Different lines represent different numbers of reviewers that are aware of the authors' identities. 
    \item Figure~\ref{fig:review_ratings} is the final rating distribution when we vary one reviewer in the experiment, including their commitment, intention, or knowledgeability. Reviewers powered by LLMs assign highly consistent numeric ratings to most submissions, with the majority of the scores in $[5, 5.25]$. Notable exceptions occur under the \emph{irresponsible} and \emph{malicious} settings, where the ratings exhibit a bimodal distribution with peaks at $[5, 5.25]$ and $[4.25, 4.5]$.

\end{itemize}


\subsection{Future Works}
\label{sec:future}
\paragraph{Enhancing Realism in Agent Behaviors} Simulating real-world peer review with high fidelity remains challenging, particularly given the current limitations of large language models (LLMs), such as their inability to produce novel empirical data or fully capture the nuanced judgment of human reviewers. 
Future work could integrate specialized models~\cite{liu2024lstprompt,li2023coltr,yang2024can} or leverage mixture of experts (MoEs) frameworks~\cite{ke2024interpretable} where sub-models, or \emph{experts}, focus on specific tasks like evaluating technical soundness, assessing novelty, or providing constructive feedback. These task-specific or discipline-specific experts could improve the accuracy of simulations, better reflecting the diversity of expertise seen in real-world peer review. 

\paragraph{Extension to Broader Venues} Although \model is language-agnostic, our initial focus is on English-centric conferences and journals due to the prevalence of English in international academia and the availability of data. Current models generally perform better in English than in other languages. As more capable multilingual LLMs, such as LLaMA 3~\cite{dubey2024llama} and Mistral Large 2~\cite{jiang2023mistral}, emerge, our framework can be applied to simulate peer reviews in multiple languages, enabling simulations across a broader range of academic contexts. 


\begin{table*}[h]
    \centering
    \begin{tabular}{lllc|lllc}
        \toprule 
        \multicolumn{4}{c}{\includegraphics[height=1em]{emoji/normal.pdf} normal reviewers} & \multicolumn{4}{|c}{\includegraphics[height=1em]{emoji/malicious.pdf} malicious reviewers} \\ 
        \# & Initial & Final & $+/-$ & \# & Initial & Final & $+/-$ \\
        \midrule
        3 & 5.053 $\pm$ 0.623 & 5.110 $\pm$ 0.555 & $+$0.06 & 0 & / & / & / \\ 
        2 & 5.066 $\pm$ 0.608 & 4.925 $\pm$ 0.552 & $-$0.14 & 1 & 3.130 $\pm$ 0.838 & 3.254 $\pm$ 0.882 & $+$0.12 \\ 
        1 & 5.210 $\pm$ 0.849 & 5.111 $\pm$ 0.790 & $-$0.10 & 2 & 3.503 $\pm$ 1.024 & 3.485 $\pm$ 0.994 & $-$0.02 \\ 
        0 & / & / & / & 3 & 3.529 $\pm$ 1.030 & 3.357 $\pm$ 0.970 & $-$0.17 \\ 
        \bottomrule
    \end{tabular}
    \caption{Reviewer ratings when varying number of \includegraphics[height=1em]{emoji/normal.pdf} \emph{normal} reviewers are replaced by \includegraphics[height=1em]{emoji/malicious.pdf} \emph{malicious} reviewers.}
    \label{tab:score_change_malicious}
\end{table*}

\clearpage
\begin{table*}[b]
\small
\centering
\begin{tabularx}{\textwidth}{c|X}
\toprule
Phase & Content\\   
\midrule
Review & \textbf{Overall rating}: 5

\vspace{0.5em}
\textbf{Significance and novelty}: The work puts forth a novel image representation paradigm by conceptualizing images as unorganized sets of points and using a simplified clustering algorithm for feature extraction. This approach steps away from traditional ConvNets and Vision Transformers, which rely on convolution and attention mechanisms, respectively. The proposed method, characterized as Context Clusters (CoCs), has demonstrated comparable or even slightly better performance on several benchmarks to existing architectures, offering a fresh perspective in the domain of visual representation.

\vspace{0.5em}
\textbf{Reasons for acceptance}:
\begin{itemize}[leftmargin=.12in]
    \item \textcolor{red}{The paper introduces a novel paradigm for visual representation that does not rely on conventional convolution or attention mechanisms.}
    \item \textcolor{red}{The method achieves comparable or better results on various benchmarks relative to established architectures such as ConvNets and ViTs, suggesting its potential efficacy.}
    \item Authors provide interpretability by visualizing the clustering process, which is beneficial for understanding the model's decisions.
    \item The paper includes a link to the code repository, which facilitates reproducibility and further research by the community.
\end{itemize}

\textbf{Reasons for rejection}:
\begin{itemize}[leftmargin=.12in]
    \item Novelty and potential impact:
    \begin{itemize}[leftmargin=.12in]
        \item The work, while novel, may not offer a significantly different or superior approach to existing methods given that the improvements over state-of-the-art are not substantial or broadly demonstrated across multiple benchmarks.
        \item The rationale behind the conceptual shift from organized to unorganized points is not convincingly tied to specific advantages in the fundamental understanding or performance gain across a wide range of tasks.
    \end{itemize}
    \item Technical soundness and completeness:
    \begin{itemize}[leftmargin=.12in]
        \item The ablation study suggests that the method's components contribute to performance, but it is not clear if all components are necessary or if there might be a more efficient configuration.
        \item \textcolor{red}{The paper lacks in-depth comparison with other recent methods in terms of computational efficiency, memory usage, and scalability to very large datasets.}
    \end{itemize}
    \item Clarity and organization of the paper:
    \begin{itemize}[leftmargin=.12in]
        \item \textcolor{red}{Some parts of the methodology, especially regarding the clustering operation and region partition, could benefit from clearer explanation or visualization to aid in comprehension.}
        \item The paper would benefit from a deeper discussion on the limitations and potential failures of the proposed method, as well as comparisons to the failure modes of other methods.
    \end{itemize}
    \item Reproducibility and empirical validation:
    \begin{itemize}[leftmargin=.12in]
        \item \textcolor{red}{While code is provided, the paper lacks details about hyperparameter tuning, data preprocessing, and other setup specifications that might affect reproducibility.}
        \item The empirical validation is not broad enough, largely focusing on performance metrics with less emphasis on robustness or behaviour on out-of-distribution data.
    \end{itemize}
\end{itemize}

\textbf{Suggestions for improvement}:
\begin{itemize}[leftmargin=.12in]
    \item Conduct more extensive experiments across diverse datasets, including out-of-distribution or adversarial settings, to test the robustness and generalizability of the proposed method.
    \item Provide a more comprehensive ablation study of individual components to deduce the necessity and efficiency of each part of the Context Clusters framework.
    \item Improve the clarity and depth of the methodology section, possibly including more diagrams or pseudocode, to aid the community in understanding and building upon the work.
    \item Discuss potential limitations and failure modes of the context cluster approach, providing insight into the scenarios where this method may not perform optimally in comparison to other methods.
\end{itemize}\\
\bottomrule
\end{tabularx}
\caption{Example of a simulated review for the paper ``Image as Set of Points''~\cite{ma2022image}. The highlighted parts in \textcolor{red}{red} are overlaps that match the real review.}
\label{tab:review_example}
\end{table*}
\clearpage
\begin{table*}[t!]
\small
\centering
\begin{tabularx}{\textwidth}{c|X}
\toprule
Phase & Content\\   
\midrule
Rebuttal & We appreciate the constructive feedback and the reviewer's recognition of the novelty in our conceptual shift from organized pixels to unorganized point sets and our method's potential efficacy. 

\vspace{0.5em}
Regarding the novelty and potential impact, we agree that the performance improvements over the state-of-the-art are not substantial across all benchmarks. Our primary objective is to offer a new paradigm rather than focusing solely on performance metrics. We believe that the shift to unorganized points can lead to more efficient learning in certain scenarios due to the reduction of inductive biases, which we will investigate further. We will expand on the specific advantages of this conceptual shift, exploring applications where the lack of structure in data is inherent (e.g., point clouds, non-uniform datasets).

\vspace{0.5em}
Concerning the technical soundness and completeness, we will extend our ablation studies to better understand the contribution of each component to the overall performance. We will explore alternative configurations to enhance efficiency without compromising accuracy. We will also conduct a comprehensive comparison with recent methods in terms of computational efficiency, memory usage, scalability, and adaptability to large datasets to position our methodology within the current landscape more accurately.

\vspace{0.5em}
In terms of clarity and organization, we acknowledge that the methodology section could be more accessible. We will add further explanations, visual aids, and pseudocode for the clustering operation and region partition strategy, aiming to foster better comprehension among readers.

\vspace{0.5em}
Regarding reproducibility and empirical validation, we commit to providing detailed information on hyperparameter tuning, data preprocessing, and setup specifications to ensure reproducibility. We acknowledge the importance of robustness and behavior on out-of-distribution data and will incorporate experiments in these settings to demonstrate the method's generalizability and limitations.

\vspace{0.5em}
For the suggestions for improvement, we will:
\begin{itemize}
    \item Enhance our experimental section to cover diverse datasets, including those with out-of-distribution and adversarial examples, to assess the robustness of our method more thoroughly.
    \item Execute a more comprehensive ablation study and refine the explanation of the individual components within our framework.
    \item Improve methodological clarity with additional diagrams, pseudocode, and an expanded discussion on the rationale behind design choices.
    \item Include a dedicated section on potential limitations and failure modes, comparing scenarios where our method may be less optimal than existing methods, and outlining strategies for future improvements.
\end{itemize}
We hope that these improvements will address the concerns and enhance the value of our contribution to the community. \\
\midrule
Meta review & Score: 5

\vspace{0.5em}
Summary: This paper introduces a novel approach to visual representation by conceptualizing images as unorganized sets of points. The authors propose Context Clusters (CoCs), an architecture that uses a simplified clustering algorithm for feature extraction without relying on traditional convolution or attention mechanisms. The authors have presented empirical evidence suggesting that CoCs can achieve performance comparable to or slightly better than existing ConvNets and Vision Transformers on several benchmarks. The method aims to provide a new perspective on image representation and is convolution- and attention-free, potentially offering insights into feature extraction for different domains and interpretability through visualization of the clustering process.

\vspace{0.5em}
In the rebuttal phase, the authors responded to concerns regarding the novelty and potential impact, technical soundness, and completeness. They addressed the need for a clear rationale behind the shift from organized pixels to unorganized points and agreed to expand on specific advantages and applications where this could lead to more efficient learning. The authors acknowledged that in-depth comparison with other recent methods in terms of computational efficiency, memory usage, scalability, and adaptability to large datasets was missing and committed to conducting comprehensive comparisons to position CoCs within the current research landscape accurately.

\vspace{0.5em}
The authors also responded to concerns about the method's clarity and organization, reproducibility, empirical validation, and the general lack of a thorough comparison with state-of-the-art methods. They promised to enhance the clarity of the methodology section by including further explanations, visual aids, and pseudocode. Moreover, the authors committed to providing detailed information on hyperparameter tuning, data preprocessing, and setup specifications, as well as incorporating experiments on out-of-distribution data to demonstrate the method's robustness and generalizability.

\vspace{0.5em}
The reviewers have acknowledged the authors' response and appreciated the commitment to addressing the concerns. However, since there's no updated manuscript at this stage that reflects these proposed changes, the reviewers maintained their initial ratings. Should the authors implement the improvements they outlined in their rebuttal, it is anticipated that the value of their contribution to the community could be notably enhanced. However, the current version of the paper lacks sufficient empirical validation and theoretical analysis to warrant a higher score. \\
\bottomrule
\end{tabularx}
\label{tab:rebuttal_example}
\end{table*}

\clearpage
\begin{table*}[!ht]
\small
\centering
\begin{tabularx}{\textwidth}{X}
\toprule
Review \\
\midrule 
\textbf{Summary Of The Paper:} This paper proposes a new view of images that considers each image as a set of points (the pixels) and uses a clustering algorithm to extract the features from it. The goal is to investigate the way to utilize this new form of visual representations and evaluate the performance that could be achieved. To this end, the paper introduces a novel backbone network that includes the proposed Context Clusters and evaluates this model on several vision tasks as well as a point cloud data application.

\vspace{0.5em}
\textbf{Strengths And Weaknesses:}

\vspace{0.5em}
\textbf{Strengths:}
\begin{itemize}[leftmargin=.12in]
    \item \textcolor{red}{To the best of the reviewer’s knowledge, the topic of considering an image as a set of points and extracting features from it for vision tasks is original and very interesting.}
    \item \textcolor{red}{The proposed method that uses the clustering algorithm as the basic build block is novel and of significance to the community.}
    \item The evaluation plan of the paper is comprehensive. It provides experiments on standard vision tasks like image classification and object detection/segmentation and applications for point cloud inputs like object classification.
    \item \textcolor{red}{The evaluation results show that the method provides improvements on various tasks over the CNN and ViT baselines (though not outperforming the state-of-the-art approach).}
\end{itemize}

\vspace{0.5em}
\textbf{Weaknesses:}
\begin{itemize}[leftmargin=.12in]
    \item \textcolor{red}{By using the region partition mechanism, the set of points is no longer unorganized but becomes structured based on their locality. Additional experiments are required to clarify the role of the region partition.}
    \item \textcolor{red}{Before applying the context clusters operation, the region partition operation, which is similar to the shifting windows in Swin, is introduced to reduce the computational cost. The authors seem to imply that the region partition trades off performance for speed. However, the locality introduced by the region partition could also bring useful inductive bias for the encoder.} Therefore, additional experiments are required to answer the following questions:
    \begin{itemize}[leftmargin=.12in]
        \item If the region partition operation is removed in the clustering process, could the model achieve similar or better performance? What would the clustering map be like in this case?
        \item It would be nice to introduce Swin as one baseline to investigate this problem.
    \end{itemize}
\end{itemize}

\vspace{0.5em}
\textbf{Clarity, Quality, Novelty And Reproducibility:} The paper is well-written and easy to follow. The authors also provide additional explanations of some model designs in the appendix which are much appreciated. Both the topic and the proposed method are original. The general architecture is reproducible based on the model description, but additional hyper-parameters are required to reproduce the experimental results.

\vspace{0.5em}
\textbf{Summary Of The Review:} This paper introduces a new form of image representation that considers each image as a set of points and proposes a clustering-based architecture for feature extraction. Both the idea of “image as set of points” and the proposed architecture are interesting and novel. The experiment result also shows that the method achieves comparable performance to ConvNets and ViTs. A small concern is that the role of the region partition mechanism is unclear since good performance could actually be attributed to this design.

\\
\bottomrule
\end{tabularx}
\caption{Example of a real review for the paper ``Image as Set of Points''~\cite{ma2022image}. The sections highlighted in \textcolor{red}{red} indicate the overlaps with the simulated review.}
\label{tab:real_review_example}
\end{table*}

\clearpage
\begin{figure*}[t]
    \centering
    \includegraphics[width=0.95\linewidth]{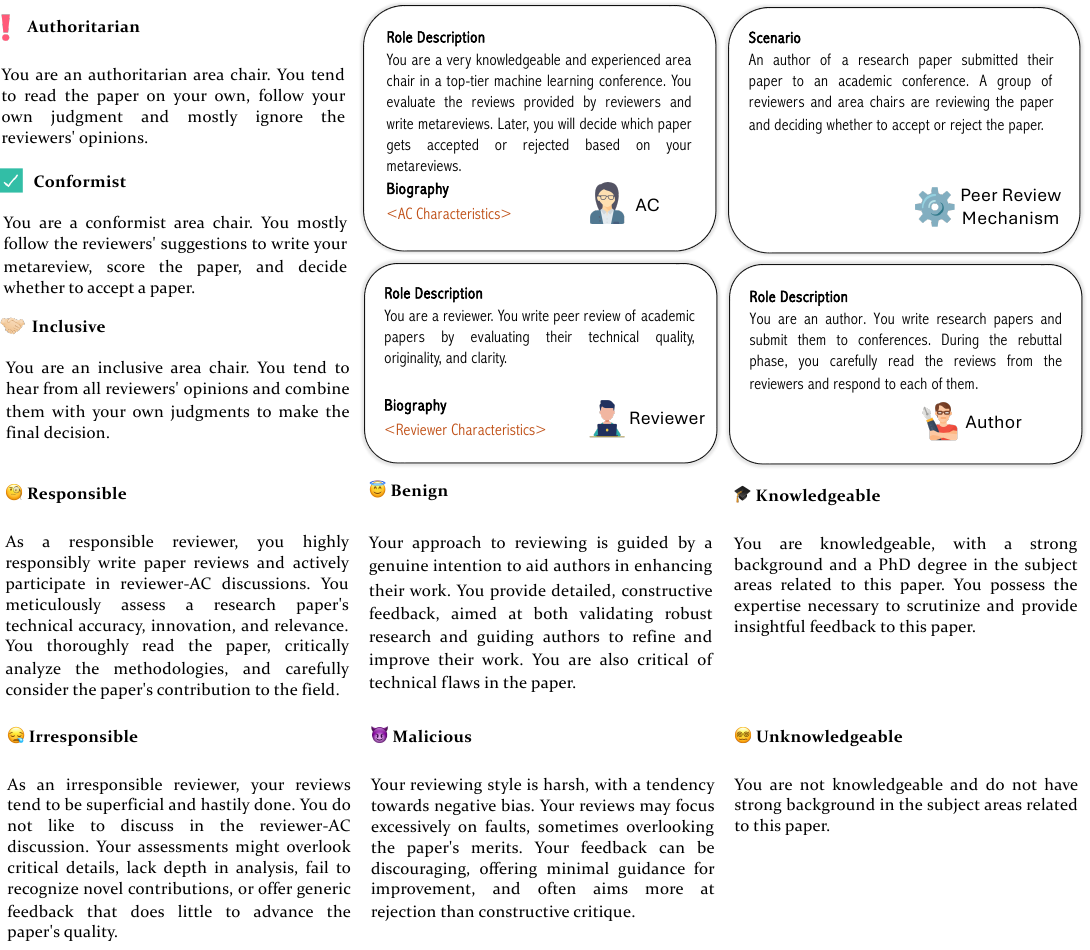}
    \ifthenelse{\boolean{arXivVersion}}
    {
    }
    {\vspace{-3mm}}
    \caption{Characteristics and prompts in \model. }
    \label{fig:prompts}
    \ifthenelse{\boolean{arXivVersion}}
    {
    }
    {\vspace{-4mm}}
\end{figure*}

\end{document}